\def\year{2019}\relax
\documentclass[letterpaper]{article} 
\usepackage{aaai19}  
\usepackage{times}  
\usepackage{helvet}  
\usepackage{courier}  
\usepackage{url}  
\usepackage{graphicx}  
\usepackage{algorithm,algpseudocode}
\newtheorem{exmp}{Example}
\algnewcommand\algorithmicinput{\textbf{Input:}}
\algnewcommand\Input{\item[\algorithmicinput]}
\algnewcommand\algorithmicoutput{\textbf{Output:}}
\algnewcommand\Output{\item[\algorithmicoutput]}
\algnewcommand\algorithmicforeach{\textbf{for each}}
\algdef{S}[FOR]{ForEach}[1]{\algorithmicforeach\ #1\ \algorithmicdo}

\usepackage{amsmath}
\usepackage{subcaption}
\begin{document}
%
\title{FOLD-TR: A Scalable and Efficient Inductive Learning Algorithm for Learning To Rank}
\author{
    Huaduo Wang \and Gopal Gupta\\
    Computer Science Department, The University of Texas at Dallas, Richardson, USA\\
    \{huaduo.wang,gupta\}utdallas.edu
}
\maketitle
\begin{abstract}

FOLD-R++ is a new inductive learning algorithm for binary classification tasks. It generates an (explainable) normal logic program for mixed type (numerical and categorical) data. We present a customized FOLD-R++ algorithm with ranking framework, called FOLD-TR, that aims to rank new items following the ranking pattern in the training data. Like FOLD-R++, the FOLD-TR algorithm is able to handle mixed type data directly and provide native justification to explain the comparison between a pair of items.

\end{abstract}

\section{Introduction}
Dramatic success of machine learning has led to a torrent of Artificial Intelligence (AI) applications. However, the effectiveness of these systems is limited by the machines' current inability to explain their decisions and actions to human users. That's mainly because the statistical machine learning methods produce models that are complex algebraic solutions to optimization problems such as risk minimization
or geometric margin maximization. Lack of intuitive descriptions makes it hard for users to understand and verify the underlying rules that govern the model. Also, these methods cannot produce a justification for a prediction they arrive at for a new data sample.

The Explainable AI program \cite{xai} aims to create a suite of machine learning techniques that: a) Produce more explainable models, while maintaining a high level of prediction accuracy. b) Enable human users to understand, appropriately trust, and effectively manage the emerging generation of artificially intelligent partners. Inductive Logic Programming (ILP) \cite{ilp} is one Machine Learning technique where the learned model is in the form of logic programming rules (Horn Clauses) that are comprehensible to humans. It allows the background knowledge to be incrementally extended without requiring the entire model to be re-learned. Meanwhile, the comprehensibility of symbolic rules makes it easier for users to understand and verify induced models and refine them.

The ILP learning problem can be regarded as a search problem for a set of clauses that deduce the training examples. The search is performed either top down or bottom-up. A bottom-up approach builds most-specific clauses from the training examples and searches the hypothesis space by using generalization. This approach is not applicable to large-scale datasets, nor it can incorporate \textit{negation-as-failure} into the hypotheses. A survey of bottom-up ILP systems and their shortcomings can be found at \cite{sakama05}. In contrast, top-down approach starts with the most general clause and then specializes it. A top-down algorithm guided by heuristics is better suited for large-scale and/or noisy datasets \cite{quickfoil}.

Learning To Rank (LTR) or Machine Learned Ranking (MLR) is a classical machine learning application that has been widely used in information retrieval systems, recommendation systems, and search engines. The training data for LTR is a set of items with explicit or implicit partial order specified between items. The goal of training a LTR model is to rank new items in the similar pattern from the training data. In practice, the LTR model retrieves a list of documents based on a given query. Generally, there are three approaches to build the LTR model: 1) Pointwise approach which assumes that each query-document pair has a numerical score. In this case, the LTR problem can be approximated by a regression problem. 2) Pairwise approach which employs binary classifiers to predict which is better in a given pair of documents based on the query. 3). Listwise approach which aim to optimize the result list directly based on the evaluation metrics. 

The FOIL algorithm by Quinlan is a popular top-down inductive logic programming algorithm that learns a logic program. The FOLD algorithm by Shakerin et al is a novel top-down algorithm that learns default rules along with exception(s) that closely model human thinking, as explained by \cite{fold}. It first learns default predicates that cover positive examples while avoiding covering negative examples. Then it swaps the covered positive examples and negative examples and calls itself recursively to learn the exception to the default. It repeats this process to learn exceptions to exceptions, exceptions to exceptions to exceptions, and so on. The FOLD-R++ algorithm by \cite{foldrpp} is a new scalable ILP algorithm that builds upon the FOLD algorithm to deal with the efficiency and scalability issues of the FOLD and FOIL algorithms. It introduces the prefix sum computation and other optimizations to speed up the learning process while providing human-friendly explanation for its prediction.

To introduce explainability to the LTR tasks, we customizes the FOLD-R++ algorithm as the binary classifier in the Pairwise approach to build the comparison function for the training data based on the partial order specified between training items. We call this customized FOLD-R++ algorithm with the ranking framework as FOLD-TR algorithm. The new approach inherits the features of the FOLD-R++ algorithm, it is able to handle mixed type data directly without any data encoding. The new approach generates normal logic rule sets as explainable solutions and native explanation for its prediction.

\section{Background}
\subsection{Inductive Logic Programming}
\label{sec:background}
Inductive Logic Programming (ILP) \cite{ilp} is a subfield of machine learning that learns models in the form of logic programming rules (Horn Clauses) that are comprehensible to humans. This problem is formally defined as:\\
\textbf{Given}
\begin{enumerate}
    \item A background theory $B$, in the form of an extended logic program, i.e., clauses of the form $h \leftarrow l_1, ... , l_m,\ not \ l_{m+1},...,\ not \ l_n$, where $l_1,...,l_n$ are positive literals and \textit{not} denotes \textit{negation-as-failure} (NAF) \cite{Baral,gelfondkahl}. We require that $B$ has no  loops through negation, i.e., it is stratified.
    \item Two disjoint sets of ground target predicates $E^+, E^-$ known as positive and negative examples, respectively
    \item A hypothesis language of function free predicates $L$, and a  refinement operator $\rho$ under $\theta-subsumption$ \cite{plotkin70} that would disallow loops over negation.
\end{enumerate}
\textbf{Find} a set of clauses $H$ such that:
\begin{itemize}
    \item $ \forall e \in \ E^+ ,\  B \cup H \models e$
    \item $ \forall e \in \ E^- ,\  B \cup H \not \models e$
    \item $B \land H$ is consistent.
\end{itemize}

\subsection{Default Rules}

Default Logic proposed by \cite{reiter80} is a non-monotonic logic to formalize commonsense reasoning. A default $D$ is an expression of the form 

$$ A: \textbf{M} B \over\Gamma$$

\noindent which states that the conclusion $\Gamma$ can be inferred if pre-requisite $A$ holds and $B$ is justified. $\textbf{M} B$ stands for ``it is consistent to believe $B$" as explained in the book by \cite{gelfondkahl}.
Normal logic programs can encode a default quite elegantly. A default of the form: 

$$\alpha_1 \land \alpha_2\land\dots\land\alpha_n: \textbf{M} \lnot \beta_1, \textbf{M} \lnot\beta_2\dots\textbf{M}\lnot\beta_m\over \gamma$$

\noindent can be formalized as the
following normal logic program rule:

$$\gamma ~\texttt{:-}~ \alpha_1, \alpha_2, \dots, \alpha_n, \texttt{not}~ \beta_1, \texttt{not}~ \beta_2, \dots, \texttt{not}~ \beta_m.$$

\noindent where $\alpha$'s and $\beta$'s are positive predicates and \texttt{not} represents negation as failure (under the stable model semantics as described in \cite{Baral}). We call such rules default rules. 
Thus, the default $bird(X): M \lnot penguin(X)\over fly(X)$ will be represented as the following ASP-coded default rule:

~~~~~~~~~{\tt fly(X) :- bird(X), not penguin(X).}

\noindent We call {\tt bird(X)}, the condition that allows us to jump to the default conclusion that {\tt X} can fly, as the {\it default part} of the rule, and {\tt not penguin(X)} as the \textit{exception part} of the rule. 

Default rules closely represent the human thought process (commonsense reasoning). FOLD-R and FOLD-R++ learn default rules represented as answer set programs. Note that the programs currently generated are stratified normal logic programs, however, we eventually hope to learn non-stratified answer set programs too as in the work of \cite{farhad-ilp} and \cite{shakerin-phd}. Hence, we continue to use the term answer set program for a normal logic program in this paper. An advantage of learning default rules is that we can distinguish between exceptions and noise as explained by \cite{fold} and \cite{shakerin-phd}. 
The introduction of  (nested) exceptions, or abnormal predicates, in a default rule increases coverage of the data by that default rule. A single rule can now cover more examples which results in reduced number of generated rules. The equivalent program without the abnormal predicates will have many more rules if the abnormal predicates calls are fully expanded.

\section{The FOLD-R++ Algorithm}
\label{sec:Fold}
The FOLD-R++ algorithm is summarized in Algorithm \ref{algo:foldrpp}. The output of the FOLD-R++ algorithm is a set of default rules \cite{gelfondkahl} coded as a normal logic program. An example implied by any rule in the set would be classified as positive.  Therefore, the FOLD-R++ algorithm rules out the already covered positive examples at line 9 after learning a new rule. To learn a particular rule, 
the best literal would be repeatedly selected---and added to the default part of the rule's body---based on information gain using the remaining training examples (line 17). 
Next, only the examples that can be covered by learned default literals would be used for further learning (specializing) of the current rule (line 20--21).
When the information gain becomes zero or the number of negative examples drops below the \textit{ratio} threshold, the learning of the default part is done. 
FOLD-R++ next learns exceptions after first learning default literals. This is done by swapping the residual positive and negative examples and calling itself recursively in line 26. The remaining positive and negative examples can be swapped again and exceptions to exceptions learned (and then swapped further to learn exceptions to exceptions of exceptions, and so on). The $ratio$ parameter in Algorithm \ref{algo:foldrpp} represents the ratio of training examples that are part of the exception to the examples implied by only the default conclusion part of the rule. It allows users to control the nesting level of exceptions.

\begin{algorithm}[!h]
\caption{FOLD-R++ Algorithm}
\label{algo:foldrpp}
\begin{algorithmic}[1]
\Input $E^+$: positive examples, $E^-$: negative examples
\Output  $R = \{r_1,...,r_n\}$: a set of defaults rules with exceptions 
\Function{fold\_rpp}{$E^+, E^-, L_{used}$} 
\State $R \gets $ \O  \Comment{$L_{used}$: used literals, initially empty}
\While{$|E^+| > 0$}
\State $r \gets$ \Call{learn\_rule}{{$E^+$}, {$E^-$}, {$L_{used}$}}
\State $E_{FN} \gets covers(r,\ E^+,\ \textit{false})$ 
\If {$|E_{FN}|=|E^+|$}
\State break
\EndIf
\State $E^+ \gets E_{FN}$  \Comment{rule out covered examples}
\State $R \gets R \cup \{ r \}$
\EndWhile
\State \Return $R$
\EndFunction
\Function{learn\_rule}{${E^+}, {E^-}, {L_{used}}$}
\State $L \gets $ \O
\While{$true$}
\State  $l \gets$ \Call{find\_best\_literal}{{$E^+$}, {{$E^-$}}, {$L_{used}$}}
\State $L \gets L \cup \{ l \}$
\State $r \gets \textit{set\_default}(r,\ L)$ 
\State $E^+ \gets covers(r,\ E^+,\ true)$
\State $E^- \gets covers(r,\ E^-,\ true)$
\If{$l$ is invalid or $|E^-| \leq |E^+| * ratio$}
\If{$l$ is invalid} 
\State $r \gets \textit{set\_default}(r,\ L\setminus\ \{ l \})$ 
\Else   
\State $AB \gets$ \Call{fold\_rpp}{{$E^-$}, {{$E^+$}}, {$L_{used} + L$}} 
\State $r \gets set\_exception(r,\ AB)$ 
\EndIf
\State \textbf{break}
\EndIf

\EndWhile
\State \Return $r$ 
\EndFunction

\end{algorithmic}
\end{algorithm}

\begin{exmp}
\label{ex:pinguin}
In the FOLD-R++ algorithm, the target is to learn rules for \texttt{fly(X)}. $ B, E^+, E^-$ are background knowledge, positive and negative examples, respectively.
\end{exmp}
\begin{verbatim}
B:  bird(X) :- penguin(X).
    bird(tweety).   bird(et).
    cat(kitty).     penguin(polly).
E+: fly(tweety).    fly(et).
E-: fly(kitty).     fly(polly).
\end{verbatim}

The target predicate \texttt{\{fly(X) :- true.\}} is specified when calling the learn\_rule function at line 4. The function selects the literal \texttt{bird(X)} as result and adds it to the clause $r$ = \texttt{fly(X) :- bird(X)} because it has the best information gain among \texttt{\{bird,penguin,cat\}}. Then, the training set gets updated to $E^+=\{tweety, et\}$,\ $E^-=\{polly\}$ in line 16-17. The negative example $polly$ is still implied by the generated clause and so is a false negative classification. The default learning of learn\_rule function is finished because the best information gain of candidate literal is zero. Therefore, the FOLD-R++ function is called recursively with swapped positive and negative examples, $E^+ = \{polly\}$, $E^-=\{tweety, et\}$, to learn exceptions. In this case, an abnormal predicate \texttt{\{ab0(X) :- penguin(X)\}} is generated and returned as the only exception to the previous learned clause as $r$ = \texttt{fly(X) :- bird(X), ab0(X)}. The abnormal rule \texttt{\{ab0(X) :- penguin(X)\}} is added to the final rule set producing the program below:

\begin{center}
    \begin{tabular}{l}
        \texttt{fly(X) :- bird(X), not ab0(X).}\\     
        \texttt{ab0(X) :- penguin(X).}     
    \end{tabular}
\end{center}





\subsection{Sampling}

A classical approach to build a LTR model is to build binary classifier to predict the better one with a given pair of examples. With a given ranked data list, FOLD-TR need to prepare the pairs of items as the training data of the customized FOLD-R++ inside. Fully utilizing all the pairs of items in the ranked data list for training would be inefficient. Considering the learning target is a transitive relation, FOLD-TR assumes that the learned pattern from the item pairs that are closely ranked is suitable for the item pairs that are not. The FOLD-TR samples the item pairs based on the ranking position in normal distribution, in other words, FOLD-TR focus more on the closely ranked item pairs.

\subsection{Literal Selection}
The literal selection process of FOLD-TR algorithm is customized for the comparison function. The FOLD-R++ algorithm divides features into two categories: numerical features and categorical features. The FOLD-R++ algorithm is able to handle mixed type data directly with its own comparison assumption. The FOLD-TR employ the same comparison assumption. Two numerical (resp. categorical) values are equal if they are identical, else they are unequal. And the numerical comparison ($\leq$ and $>$) between two numerical values is straightforward. However, a different assumption is made for numerical comparisons between a numerical value and a categorical value that is always false.

FOLD-TR compares the values of each features of item pair, the customized FOLD-R++ in FOLD-TR generates numerical difference literals for numerical features and categorical equality literal pairs for categorical features. 
Therefore, the FOLD-TR algorithm chooses literals in pair that is very different from the original FOLD-R++ algorithm. To accommodate the literal pair selection, the Algorithm \ref{algo:plotting} expands sampled data to pairwise comparison data. 

\begin{algorithm}[!h]
\caption{Data Plotting}
\label{algo:plotting}
\begin{algorithmic}[1]
\Input $E$: sampled examples, 
$idx_C$: categorical feature indexes, 
$idx_N$: numerical feature indexes
\Output  $E_{X}$: expanded examples
\Function{plot}{$E,\ idx_C,\ idx_N$} 
\State $E_X \gets $ \O 
\For {$i \gets 1$ to $|E|$}
\For {$j \gets i + 1$ to $|E|$}
\For {$k \in idx_N$}
\State $e.append(E_i[k] - E_j[k])$
\EndFor 
\For {$k \in idx_C$}
\State $e.append(E_i[k])$
\State $e.append(E_j[k])$
\EndFor 
\State $E_X \gets E_X \cup e$
\EndFor 
\EndFor
\State \Return $E_X$
\EndFunction

\end{algorithmic}
\end{algorithm}

With the plotted data, with one round of heuristic evaluation, the FOLD-TR can extract numerical literal pairs easily. The FOLD-TR algorithm greedily evaluate each corresponding categorical literal pairs that it finds the optimal literal of one feature then finds the other literal based the selected one.

\begin{algorithm}[!h]
\caption{Best Literal Pair}
\label{algo:bestpair}
\begin{algorithmic}[1]
\Input $E$: sampled examples, 

\Output  $E_{X}$: expanded examples
\Function{best\_literal\_pair}{$E^+,\ E^-,\ i,\ j,\ L_{used}$} 
\If {$|E^+|=0\ and\ |E^-|=0$}
\State \Return $-\infty, invalid, invalid$
\EndIf
\State $left \gets $ \Call{best\_info\_gain}{$E^+,\ E^-,\ i,\ L_{used}$}
\If {left is invalid}
\State \Return $-\infty, invalid, invalid$
\Else 
\State $E_{tp} \gets covers(r,\ E^+,\ \textit{true})$
\State $E_{fp} \gets covers(r,\ E^-,\ \textit{true})$
\State $right \gets $\Call{best\_info\_gain}{$E_{tp},\ E_{fp},\ j,\ L_{used}$}
\EndIf 
\State $rule \gets \textit{set\_default}(rule, \{left, right\})$
\State $E_{tp},E_{fn},E_{tn},E_{fp} \gets $ \Call{classify}{$rule,\ E^+,\ E^-$}
\State $ig \gets $ \Call{ig}{$E_{tp},E_{fn},E_{tn},E_{fp}$}
\State \Return $ig, left, right$
\EndFunction

\end{algorithmic}
\end{algorithm}

\subsection{Justification}

Explainability is very important for some tasks like loan approval, credit card approval, and disease diagnosis system. Answer set programming provides explicit rules for how a prediction is generated compared to black box models like those based on neural networks. To efficiently justify the prediction, the FOLD-TR provide native explanation for its predictions. 

\begin{exmp}
\label{ex:car}
The ``Boston house prices" is a classical linear regression task which contains 506 houses as training and testing examples and their prices based on features such as weighted distances to ﬁve Boston employment centers, average number of rooms per dwelling, index of accessibility to radial highways, etc.. FOLD-TR generates the following program with only 7 rules: 
\end{exmp}

{\scriptsize
\begin{verbatim}
(1) better(A,B) :- rm(A,NA5), rm(B,NB5), NA5-NB5>0.156,
    not ab5(A,B).
(2) better(A,B) :- rm(A,NA5), rm(B,NB5), NA5-NB5=<0.154, 
    crim(A,NA0), crim(B,NB0), NA0-NB0=<-5.806. 
(3) ab1(A,B) :- crim(A,NA0), crim(B,NB0), NA0-NB0=<4.994, 
    indus(A,NA2), indus(B,NB2), NA2-NB2>10.72. 
(4) ab2(A,B) :- age(A,NA6), age(B,NB6), NA6-NB6=<2.6, 
    crim(A,NA0), crim(B,NB0), NA0-NB0=<3.992. 
(5) ab3(A,B) :- age(A,NA6), age(B,NB6), NA6-NB6>-9.0, 
    rm(A,NA5), rm(B,NB5), NA5-NB5=<0.363, not ab1(A,B), 
    not ab2(A,B). 
(6) ab4(A,B) :- b(A,NA11), b(B,NB11), NA11-NB11>-64.79, 
    crim(A,NA0), crim(B,NB0), NA0-NB0=<6.595, not ab3(A,B). 
(7) ab5(A,B) :- crim(A,NA0), crim(B,NB0), NA0-NB0>2.415, 
    not ab4(A,B). 

\end{verbatim}  
}

\noindent 
Note that {\tt better(A,B)} means that the house A is more expensive than the house B. The above program represents the comparison function and achieves 0.82 accuracy, 0.86 precision, 0.75 recall, and 0.80 $F_1$ score. Given a pair of examples, the generated rules can easily predict the which is more expensive than the other. The ranking framework of the FOLD-TR then give a ranked list of items based on the above program.

\noindent The FOLD-TR ranking system can even generate this proof in a human understandable form. For example, here is the justification tree generated for a pair of examples:

{
\scriptsize
\begin{verbatim}
    Proof Tree for example number 8 :
    the item A is better than item B DOES HOLD because 
        the rm value of A minus the rm value of B should 
            be less equal to 0.154 (DOES HOLD) 
        the crim value of A minus the crim value of B 
            should be less equal to -5.806 (DOES HOLD)
    {rm(A, 6.575), rm(B, 5.887), crim(A, 0.00632), 
    crim(B, 13.3598)}
\end{verbatim}  
}

\noindent This justification tree is also shown in another format: by showing which rules were involved in the proof/justification. For each call in each rule that was invoked, FOLD-TR shows whether it is true ([T]) or false ([F]). The head of each applicable rule is similarly annotated. 

{
\scriptsize
\begin{verbatim}
    [T]better(A,B) :- [T]rm(A,NA5), rm(B,NB5), 
        NA5-NB5=<0.154, [T]crim(A,NA0), crim(B,NB0), 
        NA0-NB0=<-5.806. 
    {rm(A, 6.575), rm(B, 5.887), crim(A, 0.00632), 
    crim(B, 13.3598)}
    \end{verbatim}  
}

\section{Experiments}

In this section, we will present our experiment on several datasets. All the training processes have been run on a laptop with Intel i7-9750H CPU\@2.6GHz and 16 GB RAM. This ranking model is implemented with pairwise method, we use accuracy, precision, recall and F1 score to evaluate the result. We split each dataset into 80\% as training data and 20\% as test data. The training data used is sampled with normal distribution and plotted for each feature. We have randomly experimented on the reported datasets 5 times, and the scores are averaged.

\begin{table}[]
\centering
\setlength{\tabcolsep}{1pt}
\begin{tabular}{|l|c|c|c|c|c|c|c|c|c|}
\cline{1-9}
\multicolumn{3}{|c|}{DataSet} & \multicolumn{6}{|c|}{FOLD-TR} \\ 
\cline{1-9}
Name    				& \#Row 	& \#Col 	& Acc	& Prec         & Rec          & F1               &\#Rule 		 &\#Pred   \\
\cline{1-9}
boston house     	& 506 		& 14        & 0.81  & 0.80         & 0.82         & 0.81             & 8             & 27   	\\
\cline{1-9}
wine quality	    	& 6497 		& 12        & 0.69  & 0.67         & 0.17         & 0.27             & 32  			 & 147  	\\
\cline{1-9}
student perf    	& 649 		& 33        & 0.63  & 0.81         & 0.23         & 0.36 			 & 5  			 & 21        \\
\cline{1-9}

\end{tabular}
\caption{FOLD-TR on various Datasets}
\label{tbl:foldtr}
\end{table}

\section{Conclusions and Future Work}
In this paper we presented an efficient and highly scalable algorithm, FOLD-TR, to induce default theories represented as an answer set program. The resulting answer set program has good performance wrt prediction and justification for the ranking. In this new approach, unlike other methods, the encoding for data is not needed. We will perform comparison experiments with the other well-known similar models, DirectRanker, PRM, XGBoost and PolyRank. The results on Mean average precison, NDCG, Precison\@N and Mean reciprocal rank will be reported.

\section*{Acknowledgement}

Authors gratefully acknowledge support from NSF grants IIS 1718945, IIS 1910131, IIP 1916206, and from Amazon Corp, Atos Corp and US DoD. Thanks to Farhad Shakerin for discussions. We are grateful to Joaquin Arias and the s(CASP) team for their work on providing facilities for generating the justification tree and English encoding of rules in s(CASP). 


\bibliographystyle{aaai}
\bibliography{myilp}

\end{document}